\documentclass{article}

\usepackage{arxiv}

\usepackage[utf8]{inputenc} 
\usepackage[T1]{fontenc}    
\usepackage{hyperref}       
\usepackage{url}            
\usepackage{booktabs}       
\usepackage{amsfonts}       
\usepackage{nicefrac}       
\usepackage{microtype}      
\usepackage{lipsum}
\usepackage{graphicx}
\usepackage{caption}
\usepackage{subcaption}
\usepackage{float}
\usepackage{amsthm}
\usepackage{enumitem}
\usepackage{fancyhdr}
\graphicspath{ {./images/} }

\usepackage{xcolor}

\usepackage[noend,ruled,algonl]{algorithm2e}

\definecolor{purple_1}{HTML}{E4CCFF}
\definecolor{green_1}{HTML}{339955}
\definecolor{green_2}{HTML}{EBFFF2}
\definecolor{orange_1}{HTML}{FEA333}

\newcommand{\CFunction}[1]{{\textcolor{green_1}{#1}}}

\SetKwProg{Prioritize}{\CFunction{\textbf{prioritize}}}{}{end}
\SetKwProg{OrderingAlg}{node\_ordering}{}{end}

\title{Associative Knowledge Graphs for Efficient Sequence Storage and Retrieval}

\author{
 Przemysław Stokłosa \\
  Institute of Management and Information Technology\\
  Bielsko-Biała, Poland \\
  \texttt{przemyslaw.stoklosa@gmail.com} \\
   \And
 Janusz A. Starzyk \\
  University of Information Technology and Management\\
   Rzeszów, Poland\\
  \texttt{starzykj@gmail.com} \\
  \And
 Paweł Raif \\
  Silesian University of Technology\\
  Gliwice, Poland\\
  \texttt{pawel.raif@gmail.com} \\
  \And
 Adrian Horzyk \\
  AGH University of Krakow\\
  Kraków, Poland\\
  \texttt{horzyk@agh.edu.pl} \\
  \And
 Marcin Kowalik \\
  Rzeszów University of Technology\\
  Rzeszów, Poland\\
  \texttt{mkowalik@prz.edu.pl} \\
}

\fancypagestyle{firstpage}{
    \fancyhf{} 
    \fancyfoot[L]{\scriptsize © 2025. This manuscript version is made available under the 
    CC-BY-NC-ND 4.0 license (\url{http://creativecommons.org/licenses/by-nc-nd/4.0/}). 
    The final published version is available at \textit{Computer Methods and Programs in Biomedicine}, 
    \url{https://doi.org/10.1016/j.cmpb.2025.108865}.}
}

\fancypagestyle{plain}{
    \fancyhf{}
    \fancyfoot[C]{\thepage}
}

\begin{document}
\maketitle
\thispagestyle{firstpage}
\begin{abstract}
\textbf{Background and Objective:}
The paper addresses challenges in storing and retrieving sequences in contexts like anomaly detection, behavior prediction, and genetic information analysis.
Associative Knowledge Graphs (AKGs) offer a promising approach by leveraging sparse graph
structures to encode sequences.
The objective was to develop a method for sequence storage and retrieval using AKGs that maintain high memory capacity and context-based retrieval accuracy while introducing algorithms for efficient element ordering.

\textbf{Methods:}
The study utilized Sequential Structural Associative Knowledge Graphs (SSAKGs).
These graphs encode sequences as transitive tournaments with nodes representing objects and edges defining the order.
Four ordering algorithms were developed and tested:
Simple Sort, Node Ordering, Enhanced Node Ordering, and Weighted Edges Node Ordering.
The evaluation was conducted on synthetic datasets consisting of random sequences of varying lengths and distributions, and real-world datasets,
including sentence-based sequences from the NLTK library and miRNA sequences mapped
symbolically with a window-based approach.
Metrics such as precision, sensitivity, and specificity were employed to assess performance.

\textbf{Results:}
The Weighted Edges Node Ordering algorithm demonstrated superior precision and resilience to graph density.
In real-world applications, sentence retrieval achieved precision rates of 94.7\%-97.3\% for contexts of 8--10 words, while miRNA sequence retrieval
using a 6-nucleotide window reached 99.6\% precision at longer contexts.
SSAKGs exhibited quadratic growth in memory capacity relative to graph size.

\textbf{Conclusions:}
This study introduces a novel structural approach for sequence storage and retrieval.
Key advantages include no training requirements, flexible context-based reconstruction,
and high efficiency in sparse memory graphs.
With broad applications in computational neuroscience and bioinformatics,
the approach offers scalable solutions for sequence-based memory tasks.

\end{abstract}


\section{Introduction}\label{sec:introduction}

    Associative networks and memories have been a cornerstone of computational neuroscience for decades providing models for how the brain stores and retries information based on context.
    While classical Hopfield networks were limited in their pattern capacity, modern advancements such as Dense Associative Memories and Transformers have significantly expanded their capabilities.
    Classical Hopfield networks, have limitations in pattern capacity, storing about 14$\%$ of the total number of neurons.
    Modern Hopfield networks, introduced by Krotov and Hopfield in 2016~\cite{krotov_dense_2016}, improved storage capacity with a new energy function.
    Dense Associative Memories~\cite{ramsauer_hopfield_2021} further extended these capabilities.

    Transformers~\cite{Vaswani2017}, introduced later, eliminate the need for recurrent and convolutional neural networks, offering faster and more parallelizable structures.
    Hierarchical Temporal Memory (HTM), by~\cite{Hawkins2006a} draws inspiration from the human brain’s structure.
    Projects like NuPIC (nupic.torch)~\cite{Nupic,hawkins2016neurons, ahmad2019densebenefitsusinghighly} integrate HTM principles into PyTorch highlighting the benefits of sparse representations.

    Associative Long Short Term Memory LSTM~\cite{pmlr-v48-danihelka16} significantly improves the performance of LSTM in tasks requiring storing and retrieving complex data by adding special associative memory.
    Clustered Neural Network (CNN) can recall stored messages quickly and enables full parallel processing, and its modification Restricted Clustered Neural Network (R-CNN)~\cite{gripon2011sparse,DaniloRobin2015} allows to reproduce data with fewer errors.

    Knowledge Graphs offer a powerful framework for representing objects and their relationships, enabling complex decision-making~\cite{knowledge_graphs,NeoGraphData}.
    Graph Neural Networks (GNNs~\cite{Zhou2020}) has further enhanced the ability to process graph-structured data using deep learning techniques.
    The ANAKG (Active Neuro-Associative Knowledge Graphs~\cite{Horzyk2017,Horzyk2014}) utilize dynamic neurons for learning sequences within an associative knowledge graph.

    Our previous work on structural associative knowledge graphs~\cite{Starzyk2024} demonstrated the potential of this approach for storing and retrieving scenes.
    In this paper, we extend this research to sequence storage and retrieval.

    \paragraph{Our Contribution:}

    \begin{enumerate}[label=(\roman*)]
        \item{We show that the same structural principles can be effectively applied to sequences, preserving the key properties of memory size, graph density, and context requirements.}
        \item{A key challenge in applying structural approaches to sequence storage is the correct ordering of retrieved sequence elements.
        Our key contribution lies in developing efficient algorithms for ordering retrieved sequence elements.
        This paper compares the efficiency of different sequence ordering algorithms and presents testing results on various datasets.}
        \item {The paper explicitly demonstrates its methods using biomedical datasets, such as miRNA sequences and linguistic datasets that mimic patient medical records.}
        \item{The SSAKG source code, Jupyter examples, and tutorials are available on GitHub  \cite{Sto2024}.
        The implemented software is also available as a 'ssakg' package on the official Python Package Index (PyPI)
            repository (\url{pypi.org/project/ssakg}) \cite{Sto2024_ssakg_pypi}.}
    \end{enumerate}

\section{Methods}\label{sec:methods}

    \subsection{Building a Knowledge Graph}\label{subsec:building-a-knowledge-graph}

    Associative knowledge graphs were developed to rapidly construct semantic memories that store associations between observed events, actions, and objects.
    These graphs form the foundation of knowledge about the outside world, making them invaluable for autonomous learning systems in open environments.

    In this paper, we use structural associative knowledge graphs to store and retrieve sequences involving diverse objects.
    In these graphs, the structure itself is used to recreate the contents of memory based on a given context, and interconnection weights are then used to restore the proper order of the stored sequence elements.

    The structure of the knowledge graph develops automatically by progressively adding information, such as sentences or sequences of numbers.
    Synaptic connections and their strengths between represented elements are recorded during the graph's creation.
    Each node in the knowledge graph represents a specific object, word, or concept.
    As knowledge is input, these nodes become interconnected.
    However, when a large amount of information is recorded with a fixed number of nodes, synaptic connections may undergo multiple modifications, reducing or losing the resolution of the original information.

    In this work, we extensively analyze the structural properties of knowledge graphs, emphasizing the importance of knowledge graph density, measured by the ratio of used synaptic connections to the total possible synaptic connections among the nodes.
    We show that the theoretical analysis performed for scene memory~\cite{Starzyk2024} can be applied to sequence memory.
    Based on this analysis, we accurately estimate the most important dependencies between the number of nodes in the graph, graph density, context size, and the number of saved sequences.
    This allows us to determine and use the appropriate graph size or required context based on the size and number of sequences stored within the associative knowledge graph memory (referred to as graph memory).

    A crucial aspect of reproducing a sequence is the proper ordering of its elements.
    An additional challenge is that the given context does not necessarily preserve the order of the context elements in the stored sequence.
    A key contribution of this paper is demonstrating that the sparsity of the graph memory enables high memory capacity, which increases quadratically with the size of the graph.

    \paragraph{Definition 1} \textit{Sequential Structural Associative Knowledge Graph (SSAKG) is a graph in which each sequence is represented by a corresponding transitive tournament.}

    The SSAKG (referred to as a sequential graph or a graph) density becomes a crucial factor influencing memory storage capacity.
    However, the sequences stored should be sufficiently intricate to enable sequence retrieval with a relatively small context (i.e., the number of observed objects of a sequence).
    Additionally, to maintain graph sparsity, a large number of unique objects must be used in the stored sequences.

    Sequences loaded into memory are represented by the corresponding directed graph in which each vertex is connected to all vertices with higher numbers than itself.
    Such graph is called a \textbf{transitive tournament.} The transitive tournament is characterized by its hierarchical ordering and its acyclic nature where there is a single directed edge between every pair of vertices.

    Each transitive tournament is embedded within the SSAKG memory graph.
    The individual elements of the sequence are placed as edges connecting the vertices of the graph, with edges labeled by corresponding numbers.

    The order of the elements corresponds to the corresponding connections, i.e.\ the first element connects to all the others, the second to all the others except the first, and so on.

    \newtheorem{example}{Example}

    \begin{example}
        As an example, let's consider a SSAKG graph that contains 20 vertices and can store sequences consisting of numbers 1 to 20.
        Let's assume that we first store the sequence [2,6,11] in this graph.
        The first element of this sequence (no\@.\ 2) is connected via the directed edges to the numbers 6 and 11.
        The second element (no\@.\ 6) connects to no\@.\ 11.
        No edge is derived from element no\@.\ 11 (Figure~\ref{fig:graph_sequences}a).
        Figure~\ref{fig:graph_sequences}b shows the recorded sequence [2,6,11] and the sequence  [11,8,2].
        \label{example:graphs}

        \begin{figure}[H]
            \centering
            \includegraphics[width=0.9\hsize]{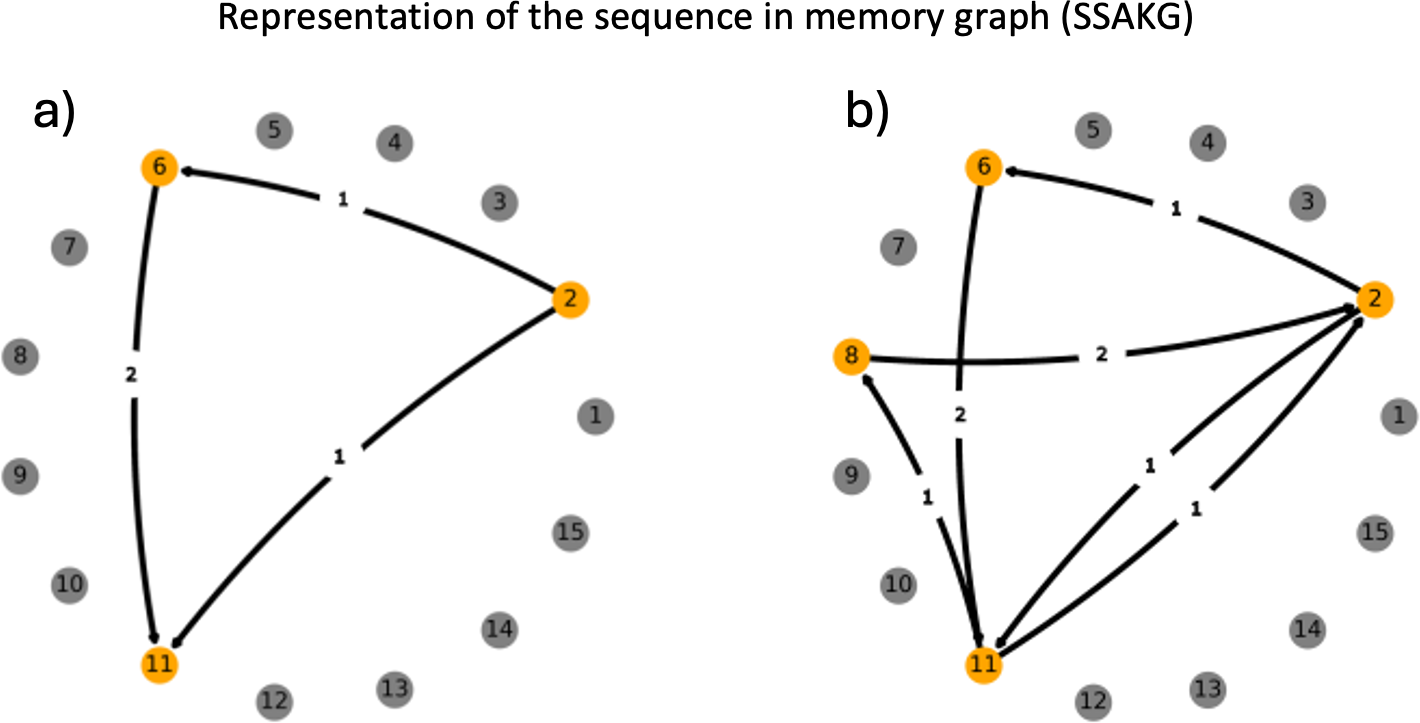}
            \caption{Sequences stored in the memory graph. Figure a) shows the stored sequence [2,6,11]. Figure b) shows the stored sequences [2,6,11] and [11,8,2].}
            \label{fig:graph_sequences}
        \end{figure}
    \end{example}

    \paragraph{Definition 2} \textit{Sequential graph \textbf{memory storage capacity} is the maximum number of sequences that can be uniquely recalled without errors.}

    Both aspects of this definition are essential.
    A recalled sequence is considered error-free if all elements of the sequence are correctly recalled.
    A recalled sequence is deemed unique if no other sequence matching the provided context is recalled.

    In the proposed approach to associative memory, where \textit{transitive tournament} subgraphs represent individual sequences, the storage capacity strongly depends on the graph's density.
    This density is influenced by the number of nodes (representing objects) and the size and quantity of stored sequences.
    As more sequences are added, the graph's density increases.
    However, once the density surpasses a critical threshold, recalling stored memories without error becomes impossible.
    Therefore, determining this \textbf{critical density} is crucial as it defines the memory capacity.

    When a knowledge graph becomes overly dense due to a limited number of unique objects, virtual objects can be a valuable solution.
    Virtual objects help maintain a sparse graph structure even when dealing with a large number of data.
    They are created by combining an object's symbol with its location in the sequence.

    This section investigates the relationship between the density of a SSAKG graph and the quantity of stored sequences.
    Additional characteristics like color, position, or size can be incorporated to further differentiate virtual objects.
    While it's theoretically possible to reconstruct a sequence using the scene memory approach described in~\cite{Starzyk2024} if all subsequent elements are converted to virtual objects, this would require context elements to also include their precise sequence location.
    To avoid this limitation, the new approach presented in this work is necessary.

    As the number of sequences grows, so does the graph density influenced by factors such as average sequence length and the total number of nodes.
    The overlapping connections between densely connected clusters (known as transitive tournament subgraphs) contribute to the overall graph density.
    To address this overlap and maintain efficiency, we must consider the effective number of synaptic connections within an SSAKG that stores multiple transitive tournament subgraphs.

    If the density \( d_{i}\) of a directed graph spanning \( n\) vertices is known, then the number of oriented edges in this graph can be calculated from the formula
    \begin{equation}
        e_{i} = d_{i}\ast n\ast\left(n-1\right)\label{eq:equation}
    \end{equation}
    By incorporating additional transitive tournament to the existing graph, we add \( e_{a}\) additional edges
    \begin{equation}
        e_{a} = (1-d_{i})\ast l_{s}\ast\left(l_{s}-1\right)\label{eq:equation2}
    \end{equation}
    where \( l_{s}\) is the length of the added sequence.
    Therefore, each new sequence written in the graph increases the graph density according to the relationship
    \begin{equation}
        d_{i+1} =\frac{e_{i}+e_{a}}{e_{n}} =  \frac{d_{i}\ast n\ast\left(n-1\right)+\left(1-d_{i}\right)\ast l_{s}\ast\left(l_{s}-1\right)}{n\ast\left(n-1\right)}\label{eq:equation3}
    \end{equation}
    Where \( e_{n}\) represents the maximum number of oriented connections in the knowledge graph spanned on \( n\) nodes.
    After simplification, we have
    \begin{equation}
        d_{i+1} =   d_{i}\ast\left(1-\frac{l_{s}\ast\left(l_{s}-1\right)}{n\ast\left(n-1\right)}\right)+\frac{l_{s}\ast\left(l_{s}-1\right)}{n\ast\left(n-1\right)} = d_{i}\ast\left(1-\xi\right)+\xi\label{eq:equation4}
    \end{equation}
    where
    \begin{equation}
        \xi  =\frac{l_{s}\ast\left(l_{s}-1\right)}{n\ast\left(n-1\right)}\label{eq:equation5}
    \end{equation}
    We can now easily connect the final density of the graph with the number of sequences \( s\) stored in it.
    Since $d_{0} = 0$, we have the final graph density:
    \begin{equation}
        d_{s} = 1-\left(1-\xi\right)^{s}\label{eq:equation6}
    \end{equation}
    and
    \begin{equation}
        s =\frac{\log\left(1-d\right)}{\log\left(1-\xi\right)}.
        \label{eq:stored_sequences}
    \end{equation}

    In equation~\ref{eq:stored_sequences} we have the interdependence of the number of stored sequences and the graph density.
    This iterative formula converges quickly assuming density \(  d = 0.5\).

    \subsection{Creating a Structured Sequence Memory}\label{subsec:creating-a-structured-sequence-memory}

    To establish a sequence memory, we can adapt the existing method used for structural scene memory.
    The key difference lies in the representation: instead of undirected complete graphs, represented by a full square matrix, scenes are now articulated as upper triangular matrices.
    This results in an asymmetric matrix that signifies a directed knowledge graph.

    To reconstruct sequences based on a given context, we can follow a similar approach to recreating scenes in structured memories.
    This involves adding an asymmetric knowledge graph matrix and its transpose to create a symmetric matrix.
    Within this matrix, sequence elements (vertices) are selected in a manner analogous to finding scene elements based on context.

    Ultimately, the maximum density of a knowledge graph representing sequential memory will be similar to that of a scene memory.

    In a directed graph, multiple sequences can be stored and retrieved based on context.
    As more sequences are added, the graph's density increases.
    We can estimate this density using a method similar to the one used for structured scene memories~\cite{Starzyk2024}.

    It is important to note that adding a directed graph matrix and its transpose can potentially double the density of the resulting symmetric matrix.
    Therefore, the maximum density of the knowledge graph used for recording sequences might be limited to half that of scene memory.
    However, this effect is offset by the fact that triangular matrices are sufficient for recording sequences, while square matrices with nearly twice as many non-zero elements are needed for scenes.
    Ultimately, the maximum density of a knowledge graph representing sequential memory will be similar to that of scene memory.

    \subsection{Sequence Elements Ordering Algorithms}\label{subsec:sequence-elements-ordering-algorithms}
    To recover a sequence from an associative knowledge graph the retrieved elements must be ordered.
    This is particularly important for large datasets, as it allows us to convert the retrieved data into a properly ordered \("\)transitive tournament graph matrix.\("\) While a simple sorting method might suffice for small, sparse graphs, more sophisticated techniques are needed for denser graphs containing elements from multiple sequences.
    This section explores four algorithms for sequence element ordering:
    \begin{enumerate}
        \item Simple Sort
        \item Node Ordering
        \item Enhanced Node Ordering
        \item Weighted Edges Node Ordering.
    \end{enumerate}

    \paragraph{Simple Sort}

    This baseline algorithm sorts the retrieved data (represented by rows in matrix M) based on the number of non-zero elements.
    It does not remove elements that do not belong to the target sequence and serves as a reference point for comparison.

    \paragraph{Node Ordering}
    This algorithm aims to generate all possible orderings of rows and columns in matrix M to obtain an upper triangular matrix (without a diagonal) containing only ones in the upper right corner.
    It prioritizes rows with n-1 non-zero elements and removes both the identified row and its corresponding column.
    This process is repeated for all such rows.
    If rows have the same number of non-zero elements, all possible ordering combinations are considered.
    A key difference from Simple Sort is the removal of additional elements from potentially different sequences.
    For instance, when selecting the first row with n-1 non-zero elements,
    the entire corresponding column is removed, potentially containing elements from other sequences.
    The pseudocode of algorithm is shown in Algorithm~\ref{alg:node_ordering}.
    To illustrate the results of the Node Ordering algorithm, let's examine a straightforward example.

    \begin{example}
        Consider $5x5$ matrix M and assume that the path obtained by the Node Ordering algorithm consists of indices $[4, 3, 3, 2, 1]$.
        These indices represent the progressively eliminated rows (starting from row 4) of successively smaller matrices.
        Initially, with the original numbering $[1, 2, 3, 4, 5]$, we arrive at smaller and smaller matrices, which align with the original row numbers as illustrated here: $[1, 2, 3, 5]\to[1, 2, 5]\to[1, 2]\to[1]$.
        This path corresponds to the desired sequence elements renumbering: $[4, 3, 5, 2, 1]$.
        \label{example:node_ordering}
    \end{example}
    It's important to note that this procedure is reversible because, given the renumbered rows, we can determine the path generated by the Node Ordering algorithm.
    The pseudocode of the Node Ordering is shown in Algorithm~\ref{alg:node_ordering},
    Algorithm~\ref{alg:prioritize_functions} shows the structure of Prioritize functions that are different for each algorithm.

    \begin{algorithm}[H]
    \scriptsize
    \DontPrintSemicolon
    \caption{Pseudocode of the Node Ordering.}
    \OrderingAlg{$(M, path, set\_of\_paths$)}{
        \KwIn{M - unordered adjacency matrix of sequence}
        \KwIn{path - current path}
        \KwIn{set\_of\_paths - set of founded paths}
        \KwOut{set of founded paths}
        n $\gets$ dimension(M)\;
        \If{$n>0$}{
            \tcp{In ideal case there is one row}
            non\_zeros\_ $\gets$ find rows contain $(n-1)$ non-zeros elements\;
            \tcp{Prioritize function is different for each algorithm}
            prioritized\_indices $\gets$ \textbf{\CFunction{prioritize}}(M,non\_zeros\_indices)\;
            \ForEach{i in prioritized\_indices}{
                path.append(i)\;
                \tcp{The algorithm has found the prioritized rows in the matrix. We can look for the next ones.}
                M $\gets$ remove row i and column i from matrix M.\;
                \textbf{node\_ordering}$(M, path, set\_of\_paths$)\;
            }
        }
        \Else{
            set\_of\_paths.append(path)\;
            \Return set\_of\_paths \;
        }

    }
    \;
    \label{alg:node_ordering}
\end{algorithm}

    \paragraph{Enhanced Node Ordering}

    This approach builds upon the Node Ordering Algorithm by considering the weights associated with connections between individual vertices of the directed graph.
    Ideally, the Node Ordering algorithm should identify only one row with n-1 number of non-zero elements.
    However, as the density of the graph increases, more and more additional elements from other sequences appear in the rows, making it more likely to find rows with the same number of non-zero elements.
    In such cases, the algorithm prioritizes edges with higher weights, reducing the number of alternative orderings.

    \paragraph{Weighted Edges Node Ordering}
    This method leverages weights directly embedded in the transitive tournament graph matrix (presented in Example~\ref{example:graphs}).
    By multiplying the individual rows of the graph matrix by successive numbers, the algorithm can find rows with appropriate weight values.

    \begin{algorithm}[H]
    \scriptsize
    \DontPrintSemicolon
    \caption{Prioritize functions.}
    \tcp{Inputs and outputs for all prioritize functions}
    \vspace{0.1cm}
    \KwIn{M - unordered adjacency matrix of sequence}
    \KwIn{non\_zeros\_indices - set of selected indices}
    \KwOut{prioritized indices}
    \;
    \tcp{Node ordering - empty function}
    \Prioritize{$(M,non\_zeros\_indices$)}{
        prioritized\_indices $\gets$ non\_zeros\_indices\;
        \Return $prioritized\_indices$
        \label{alg:prioritize_node_ordering}
    }
    \;

    \tcp{Enchanced node ordering}
    \Prioritize{$(M,non\_zeros\_indices)$}{
        Calculate weights: $w_{ij}=\frac{1}{M_{ij}}$ \;
        Summarize weights in each row \;
        prioritized\_indices $\gets$ find rows with min weights \;
        \Return prioritized\_indices
    }
    \;
    \tcp{Weighted edges node ordering}
    \Prioritize{$(M,non\_zeros\_indices)$}{
        \tcp{Determine the prioritised number, in this case the number is equal to the dimension of M.}
        prioritized\_number $\gets$  dimension(M)\;
        prioritized\_indices $\gets$ find rows with majority of prioritized numbers \;
        \Return prioritized\_indices
    }
    \;
    \label{alg:prioritize_functions}
\end{algorithm}

\section{Results}\label{sec:results}

    \subsection{Comparison of Graph Ordering Algorithms Performance}\label{subsec:comparison-of-graph-ordering-algorithms-performance}

\begin{figure}[H]
        \centering
        \includegraphics[width=\hsize]{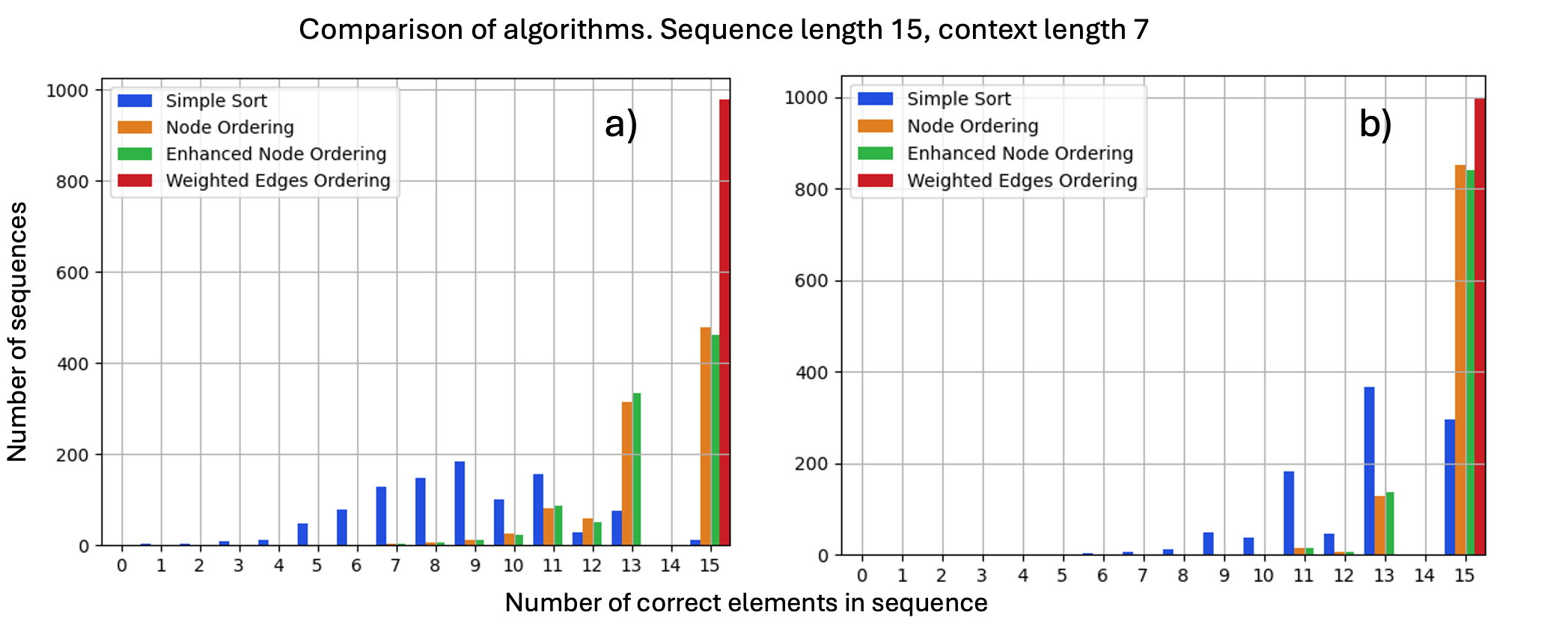}
        \caption{Different SSAGK graph sizes used based on a) 1000 elements, b) 2000 elements. The 1000 sequences were memorized. As you add repeated sequence elements, the size of the graph matrix increases.}
        \label{fig:algorithm_comparison}
    \end{figure}

    The following example uses randomly generated data as test sequences.
    When generating the example sequences, we used a uniform distribution.
    Figure~\ref{fig:algorithm_comparison} compares the performance of all four algorithms in reconstructing sequences.
    The test involved 1000 sequences, each containing 15 elements.
    Figure 4a shows the results for a graph containing 1000 elements, while Figure 4b presents the results for a 2000-element graph.

    The tests were performed on an Apple Mac Mini M1 with 16GB RAM. Depending on the complexity of the test, the duration of a single test ranged from 1 second to 1 minute.

    In both cases, the context used to reconstruct the sequence included 7 unordered elements.
    Histograms illustrate the reproducibility of sequence elements.
    The x-axis represents the number of correctly reproduced elements, with 15 indicating perfect reconstruction.
    As evident, the Weighted Edges Ordering algorithm outperforms all others.

\begin{figure}[H]
        \centering
        \includegraphics[width=0.95\hsize]{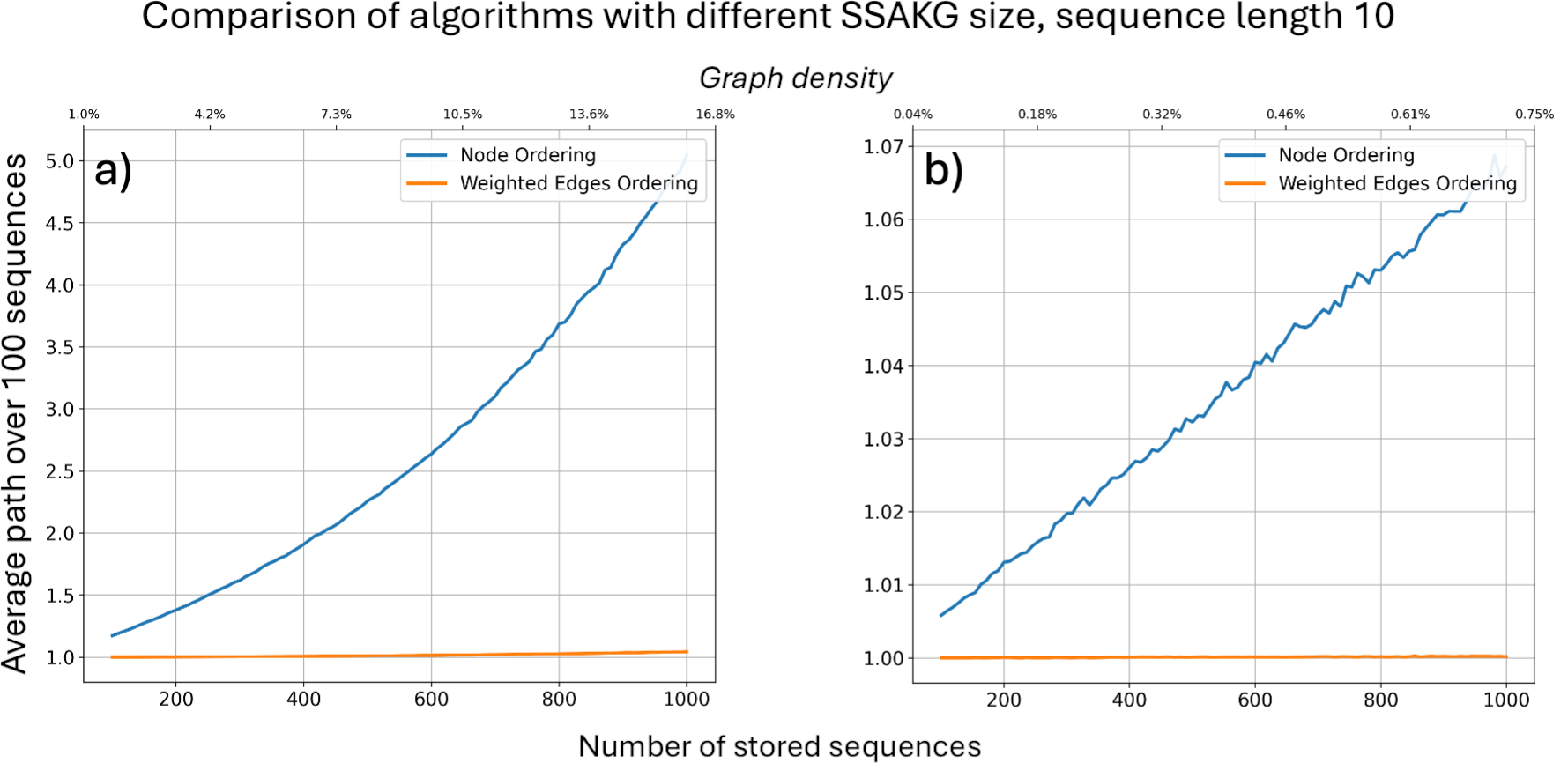}
        \hspace{0.25 in}
        \caption{Comparison of path sorting algorithms. Different graph memory (SSAKG) dimension was used a) 500, b) 2500. The sequence length is 10. The image shows results averaged over 1,000 different memory graphs.}
        \label{fig:memory_comparison}
    \end{figure}

    Figure~\ref{fig:memory_comparison} explores the average number of node orderings obtained for sequences stored in SSAKG graphs of varying sizes (all sequences were 10 elements long) assuming that all sequence elements are known.

    As expected, the average number of orderings decreases with increasing graph size (represented on the lower x-axis) and increasing graph density (upper x-axis).
    This experiment focuses on Algorithms 2 (Node Ordering) and 4 (Weighted Edges Node Ordering) since Simple Sort performed significantly worse, and Enhanced Node Ordering offered similar results to Node Ordering.
    The figure demonstrates that the Weighted Edges Node Ordering algorithm exhibits a much slower growth in the average number of node orderings, indicating its superior efficiency in accurately reconstructing sequences stored in the SSAKG graph.

    Overall, these results highlight the importance of effective sequence node ordering algorithms for accurate data retrieval from associative knowledge graphs.
    The Weighted Edges Node Ordering algorithm demonstrates superior performance compared to simpler approaches.
    This is essential for tasks like ordering genetic sequences (e.g., miRNA) or reconstructing medical event timelines.

    Computational complexity. The space complexity exhibits an asymptotic growth rate of $O(n^2)$ (where n is the number of all available symbols) and is independent of the length of the sequences (l parameter). This is because the complement matrix's space requirement depends solely on the number of symbols (n) and remains constant, irrespective of the sequence length (l) or the number of stored sequences. Regarding time complexity, the asymptotic growth is $O(n)$. While the dependence on the sequence length is challenging to estimate precisely, it is approximately $O(l^2)$.

    The number of stored sequences $n$ does not affect the read speed visibly. However, the main issue is the length of the sequence $l$. The density $d$ of the graph increases proportionally to the square of the length of the sequence $l$.

    Regarding computational requirements, SSAKG can handle numbers of over 100,000 symbols (tested on an NVIDIA DGX A100 computational server).

    \subsection{Practical Examples: Encoding Sentences in SSAKG}\label{subsec:practical-examples}

    This section presents examples based on memorizing and recalling sentences from SSAKG memory.
    The program uses the NLTK library - Natural Language Toolkit to obtain correct sentences.  \cite{bird2009natural}.
    This library enables reading sentences and words with appropriate grammar for the specified language.
    Additionally, it contains a vast collection of texts grouped into \("\)corpora\("\).
    For our experiments, we utilized a corpus based on Gutenberg’s project~\cite{Gutenberg2024}, which includes several selected book entries.
    We randomly selected several thousand sentences with interesting parameters, filtering out common words (stop words) and punctuation marks~\cite{Sarica2021}.
    To tokenise and remove stop words, we used the appropriate tools provided by the NLTK library.

    \begin{example}
        \textbf{ Sentences of Equal Length (15 words).}
        We selected 1000 random sentences, each containing 15 words.
        These sentences were prepared using the previously described method.
        The random text contained 4453 different words.
        The word distributions in the text are depicted in Figure~\ref{fig:words_15}.
        The selected sentences were then encoded into SSAKG\@.

        \begin{figure}[H]
            \centering
            \includegraphics[width=\hsize]{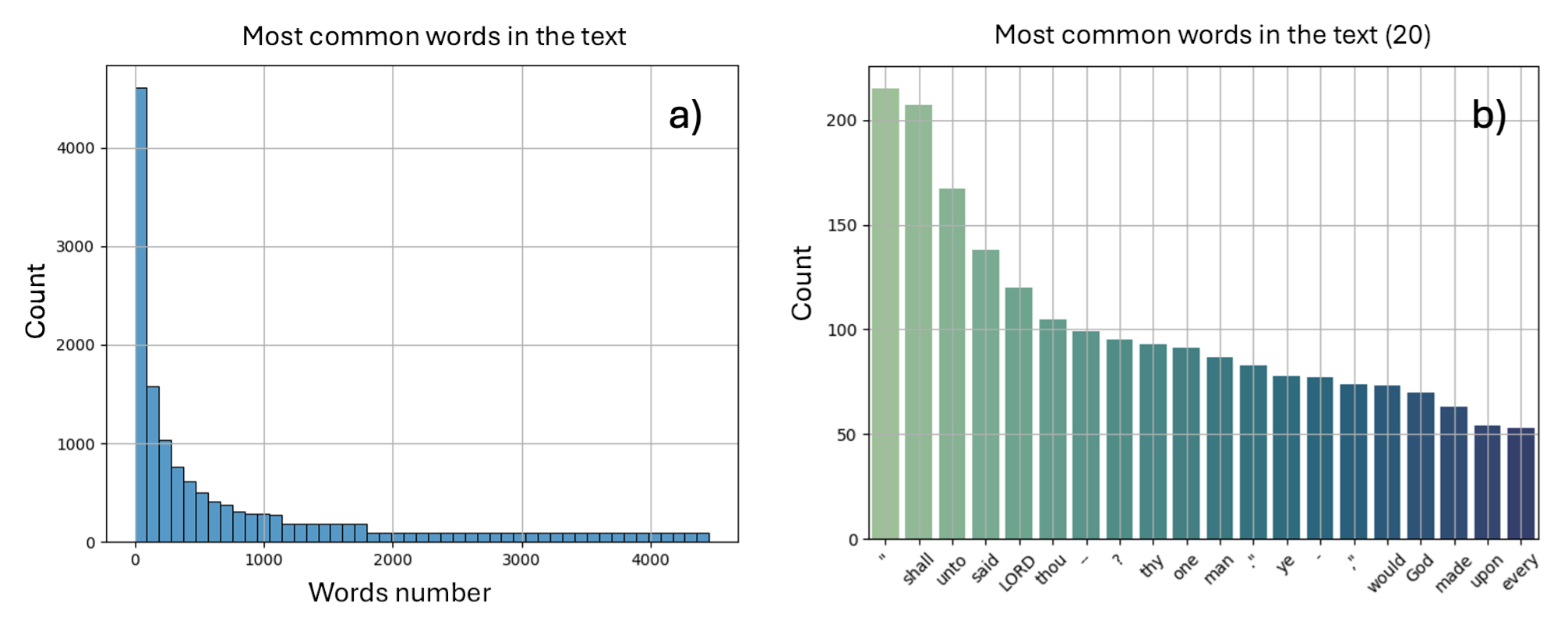}
            \caption{Words occurring in the text a) distribution of all words in the text, b) distribution of the most frequent words.}
            \label{fig:words_15}
        \end{figure}

        Subsequently, using contexts of length 8, 9, and 10, we attempted to recall these sentences from memory.
        The results are presented in Table~\ref{tab:corretness_15_words}.
        For randomly generated sequences with a flat distribution, the program successfully reproduced all sequences.

        \begin{table}[H]
            \caption{Correctness of Reading Sentences with 15 Words}
            \centering
            \begin{tabular}{ccc}
                \toprule
                Context Length & Correct Sets of Words & Correct Order \\
                \midrule
                8              & 95.1\%                & 96.3\%        \\
                9              & 96.6\%                & 96.1\%        \\
                10             & 97.3\%                & 95.9\%        \\
                \bottomrule
            \end{tabular}
            \label{tab:corretness_15_words}
        \end{table}
        \label{example_words_15}
    \end{example}

    \begin{example}
        \textbf{ Sentences of Varying Lengths (10-15 words).}
        We selected 1000 random sentences with lengths ranging from 10 to 15 words.
        The sentences were prepared using the previously described method.
        This random text contained 3037 different words.
        The word distributions in the text are depicted in Figure~\ref{fig:words_10_15}.
        The selected sentences are then encoded into SSAKG. Subsequently, using contexts of lengths 8,9, and 10, we attempted to recall these sentences from memory.
        \begin{figure}[H]
            \centering
            \includegraphics[width=\hsize]{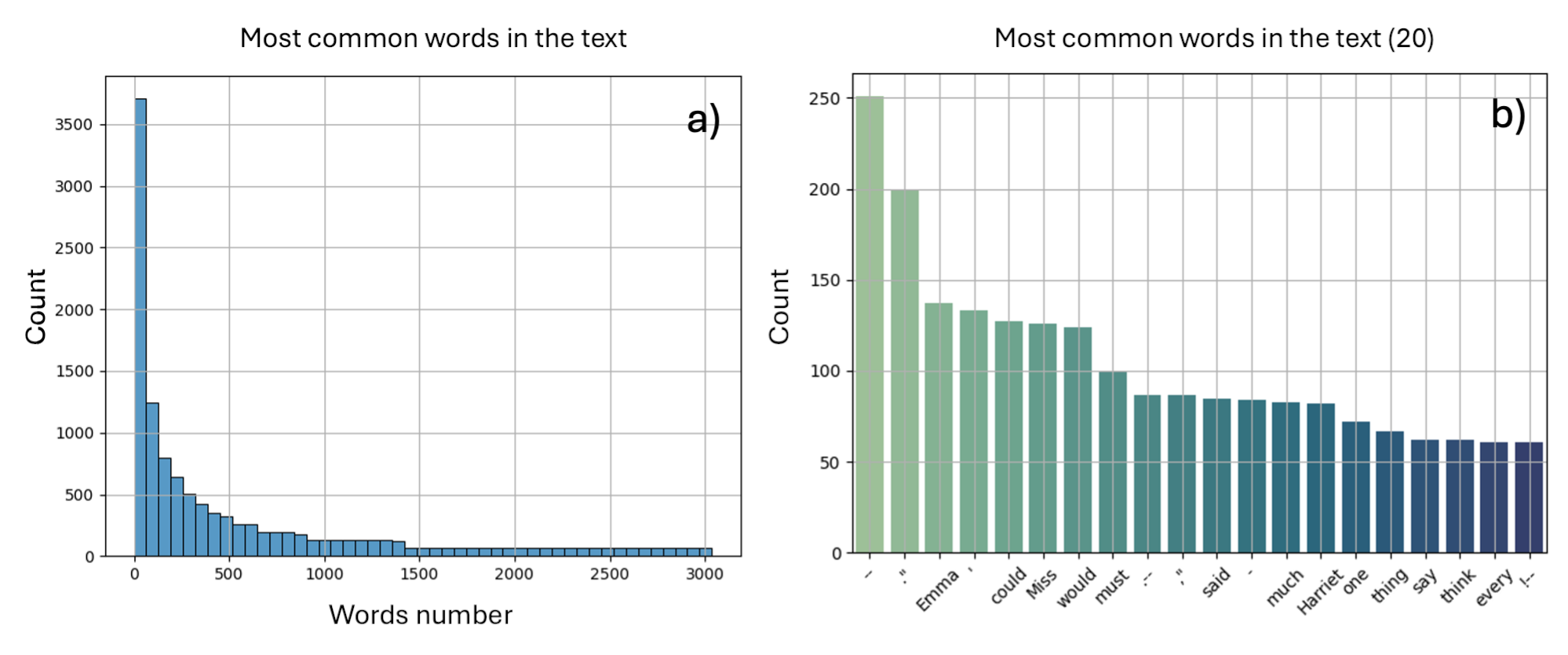}
            \caption{Words occurring in the text a) distribution of all words in the text, b) distribution of the most frequent words.}
            \label{fig:words_10_15}
        \end{figure}
        The results are presented in Table~\ref{tab:corretness_10_15_words}.
        For randomly generated sequences with a flat distribution, the program successfully reproduced all sequences.

        \begin{table}[H]
            \caption{Correctness of Reading Sentences with 10--15 Words}
            \centering
            \begin{tabular}{ccc}
                \toprule
                Context Length & Correct Sets of Words & Correct Order \\
                \midrule
                8              & 94.7\%                & 95.4\%        \\
                9              & 95.2\%                & 94.7\%        \\
                10             & 97.0\%                & 94.7\%        \\
                \bottomrule
            \end{tabular}
            \label{tab:corretness_10_15_words}
        \end{table}
        \label{example_words_10_15}
    \end{example}

    \subsection{Practical Examples:  Encoding MicroRNA Sequences in SSAKG}\label{subsec:practical-examples-mirna}
    MicroRNAs (miRNAs) are small, non-coding RNAs that
    regulate gene expression by binding to target mRNA molecules.
    While computational methods can predict miRNA-mRNA interactions, inconsistencies
    between databases remain a challenge.
    Although multiple online databases exist, they
    often report inconsistent results regarding miRNA targets, as target sequences
    can differ between databases~\cite{Videti_Paska_2022,Paraskevopoulou_2013}.
    Therefore, new and improved methods for matching miRNAs to mRNA sequences are continuously being developed.
    Our approach efficiently identifies miRNA
    sequences, focusing on their unique characteristics.
    This approach can also be adapted for matching miRNAs with mRNA targets.

\begin{figure}[h]
        \centering
        \includegraphics[width=0.5\hsize]{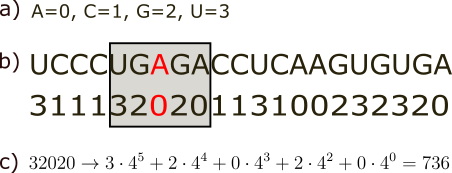}
        \caption{The mapping algorithm for converting nucleotide sequences into symbol sequences.
            (a) Nucleotide symbols. (b) Construction of corresponding window.
            (c) Calculating a symbol from a quadruplex system}
        \label{fig:mirna_encoding}
    \end{figure}

    \begin{example}
        We selected a dataset of 1000 random miRNA sequences each 20 nucleotide long.
        The MicroRNA data was downloaded from miRbase~\cite{mirbase_2010}.
        The challenge lies in representing these sequences efficiently in an SSAKG\@.
        A straightforward approach would be to represent each nucleotide as a node in the graph.
        However, this would lead to a limited number of unique symbols (four, corresponding to the four nucleotide
        bases: A, C, G, T). To increase the diversity of symbols and capture more
        complex sequence patterns, we introduced a window-based approach.
        Symbols are created based on a given nucleotide and the nucleotides surrounding it (Figure~\ref{fig:mirna_encoding}).
        Each nucleotide is numbered as shown in Figure~\ref{fig:mirna_encoding} (a).
        The sequence of numbers within the window shown in Figure~\ref{fig:mirna_encoding} (b) is then converted
        into a single symbol using a quadruplex number system Figure~\ref{fig:mirna_encoding} (c).
        This process effectively multiplies the number of possible symbols and allows for more
        nuanced representation of sequence information in the SSAKG\@.
        \label{example_mirna}
    \end{example}

    The results are presented in Table~\ref{tab:corretness_20_nucleotydes}.
    A window containing 6 nucleotides was selected in the test.
    By using this approach, we were able to significantly increase the diversity of
    symbols in the SSAKG, enabling more efficient storage and retrieval of sequence information.
    For miRNA test we created a context in with
    consecutive elements, as opposed to the standard context, which contains
    unsorted elements.
    Increasing the number of sequences in a
    dataset leads to a decrease in the accuracy of sequence reconstruction from
    context, which is due to the lack of uniqueness of context in this larger dataset.

    \begin{table}[H]
        \caption{Correctness of Reading miRNA sequences with 20 Nucleotides}
        \centering
        \begin{tabular}{ccc}
            \toprule
            Context Length & Correct Sets of Nucleotides & Correct Order \\
            \midrule
            3              & 64.10\%                     & 94.54\%       \\
            4              & 79.40\%                     & 95.09\%       \\
            5              & 86.60\%                     & 94.80\%       \\
            6              & 91.40\%                     & 94.97\%       \\
            9              & 95.70\%                     & 94.67\%       \\
            \bottomrule
        \end{tabular}
        \label{tab:corretness_20_nucleotydes}
    \end{table}

    \begin{example}

        We selected a dataset containing more than 8,000 miRNA sequences of 21 nucleotides each.
        These are all sequences of this length currently available on miRBase.
        A window containing 7 nucleotides was selected in the test.
        The results are presented in Table~\ref{tab:corretness_21_nucleotydes}.
        \begin{table}[H]
            \caption{Correctness of Reading miRNA sequences with 21 Nucleotides}
            \centering
            \begin{tabular}{ccc}
                \toprule
                Context Length & Correct Sets of Nucleotides & Correct Order \\
                \midrule
                5              & 63.04\%                     & 82.25\%       \\
                6              & 74.30\%                     & 82.37\%       \\
                9              & 84.54\%                     & 81.73\%       \\
                \bottomrule
            \end{tabular}
            \label{tab:corretness_21_nucleotydes}
        \end{table}

    \end{example}

    The paper introduces a window-based symbolic
    representation for miRNA sequences, increasing the diversity of symbols used in the graph.
    This method significantly improves storage and retrieval efficiency for nucleotide sequences, crucial for
    genomic studies and disease research.
    Precision values for retrieving miRNA sequences reached 95.7\% with a 9-context length.

    \section{Discussion and conclusion}\label{sec:discussion}

    This paper introduces a novel structural approach for storing and retrieving sequences based on a given context.
    We employed associative knowledge graphs composed of transitive tournaments that represent stored sequences.
    Our research demonstrates that these graphs can effectively create associative sequential memories by utilizing structural information about synaptic connections and integer weights representing sequence element order.

    We found that the memory capacity is determined by the size of the knowledge graph and the density of its synaptic connections.
    Our theoretical analysis reveals that memory capacity grows quadratically with the number of neurons used to build the knowledge graph.
    While smaller graphs may not yield significant sequence capacity, larger graphs can store a considerable number of sequences, allowing each node to represent multiple sequences.

    If the stored sequences have many elements in common, the result of resolving them depends on the context used. If the context is common to several stored sequences, then the algorithm will return all elements contained in several sequences. Such an extended sequence cannot be sorted properly. Such a result is considered incorrect. In our tests, we only considered it a success if the entire sequence is returned without error.

    The method of MicroRNA (miRNA) coding described in Example~\ref{example_mirna} minimizes this behavior of the algorithm. In all MicroRNA tests, sequences were selected from the database at random.

    A crucial aspect of our work was the development of algorithms for ordering sequence elements from the retrieved set of graph nodes activated by the input context.
    This paper presents results comparing the performance of these algorithms.

    We conducted memory tests using both randomly generated synthetic data and real-world datasets, such as a collection of sequences of words featuring different sentences.
    Additionally, we tested an example using a genetic database.
    These tests confirmed the validity of our findings.

    Our approach offers a novel way to store sequences efficiently.
    By leveraging the sparse nature of knowledge graphs and the relatively small size of recorded sequences, we can achieve significant memory capacity for sequences.
    Unlike other neural networks used for sequence storage, our method has several advantages:

    \begin{itemize}
        \item No training is required: The associative graph structure is created automatically as new sequences are added.
        \item Flexible context: The context used to search for and reconstruct sequences can be presented in any order.
        \item Efficient retrieval: Even a relatively small context is sufficient to reconstruct the entire sequence.
        \item The authors provide an open-source implementation of SSAKGs via the SSAKG Python package, along with Jupiter notebooks and tutorials.
        \item This accessibility promotes adoption and further exploration of the methodology in both academic and clinical settings.
        \item Biomedical researchers can adapt these techniques for anomaly detection, disease progression modeling, or personalized medicine.
    \end{itemize}

    Compared to traditional Hopfield networks, which rely on a fixed number of neurons to store memories, this method offers a more efficient solution.
    By utilizing sparse knowledge graphs and contextual connections, it can store and recall a greater number of scenes without sacrificing performance.

    The method's limitations stem from the theoretical constraints expressed in equations~\ref{eq:equation6} and~\ref{eq:stored_sequences}.
    The memory capacity, which is largely influenced by the density of the knowledge graph and the length of memorized sequences, is significantly reduced as the sequences grow longer.
    This limitation is directly linked to the parameter $\xi$ in equation~\ref{eq:equation5}, which increases with increasing sequence length.
    This approach has broad applications, from computational neuroscience to bioinformatics, offering scalable solutions for sequence-based memory tasks.

    Our future work related to this topic will focus on two key aspects.
    First, we aim to compare the proposed structural associative memory with contemporary associative memory solutions, such as Modern Hopfield Networks (MHN), LSTM, and NuPIC.
    A comparative analysis of these two solutions will be the subject of our upcoming article.
    Second, we plan to enhance the presented node ordering concept by incorporating novel elements like using prime numbers to represent sequence order and micro-columns to increase memory capacity.

    \section*{Credit authorship contribution statement}
    \textbf{Przemysław Stokłosa:} Software design, writing, Methodology, Conceptualization, Writing – review \& editing,
    Project administration, Data acquisition, Resources.
    \textbf{Janusz A. Starzyk:} Methodology, Conceptualization, Software writing, Writing – review \& editing, Data acquisition.
    \textbf{Paweł Raif:} Writing – review \& editing, Resources, Project administration, Software tests, Funding acquisition.
    \textbf{Adrian Horzyk:} Writing – review \& editing, Resources.
    \textbf{Marcin Kowalik:} Writing – review \& editing, Data acquisition, Resources.
    \section*{Supplementary materials}
    \begin{itemize}
        \item Source code available: \url{https://github.com/PrzemyslawStok/ssakg}
        \item Python package \textbf{ssakg} available.
        For the installation process, use the pip package manager: \emph{pip install ssakg}.
    \end{itemize}
    \section*{Declaration of competing interest}
    The authors declare that they have no known competing financial interests
    or personal relationships that could have appeared to influence the work
    reported in this paper.

\bibliographystyle{unsrt}
\bibliography{bibliography-bibtex.bib}

\end{document}